\documentclass[
twocolumn,
]{main}

\usepackage{listings}
\usepackage{latexsym}
\usepackage{amssymb}
\usepackage{amsmath}
\usepackage{amsthm}
\usepackage{booktabs}
\usepackage{enumitem}
\usepackage{graphicx}
\usepackage{color}
\usepackage{algorithm}
\usepackage{algorithmicx}
\usepackage{algpseudocode}
\usepackage{hyperref}
\begin{document}

\copyrightyear{2025}
\copyrightclause{Copyright for this paper by its authors.
Use permitted under Creative Commons License Attribution 4.0 International (CC BY 4.0).}


\title{A DbC Inspired Neurosymbolic Layer for Trustworthy Agent Design}


\author[A]{Claudiu Leoveanu-Condrei}[
email=leo@extensity.ai,
url=https://futurisold.github.io/,
]
\address[A]{ExtensityAI, 20 Stefan-Fadinger-Straße, Wels, 4600, Austria}





\begin{abstract}
Generative models, particularly Large Language Models (LLMs), produce fluent outputs yet lack verifiable guarantees. We adapt Design by Contract (DbC) and type-theoretic principles to introduce a contract layer that mediates every LLM call. Contracts stipulate semantic and type requirements on inputs and outputs, coupled with probabilistic remediation to steer generation toward compliance. The layer exposes the dual view of LLMs as semantic parsers and probabilistic black-box components. Contract satisfaction is probabilistic and semantic validation is operationally defined through programmer-specified conditions on well-typed data structures. More broadly, this work postulates that any two agents satisfying the same contracts are \emph{functionally equivalent} with respect to those contracts.
\end{abstract}

\begin{keywords}
    neurosymbolic AI \sep
    design-by-contract \sep
    probabilistic programming \sep
    large language models
\end{keywords}

\maketitle

\section{Introduction}
The increasing integration of generative models, especially Large Language Models (LLMs), into diverse applications necessitates formal approaches to ensure the dependability of their outputs \citep{yan2024large, huang2024survey}. LLMs are susceptible to producing outputs that, despite being syntactically plausible, may be factually incorrect or semantically misaligned with user expectations \citep{chen2021evaluating, deng2024don, zhang2023safetybench}.

Standard software engineering paradigms like Design by Contract (DbC) \citep{meyer1997object} offer principles for constructing reliable systems by enforcing explicit specifications, termed \emph{contracts}, that detail the mutual obligations between software components. This research extends and adapts these principles to the inherently probabilistic and semantic operational domain of modern generative models.

\citet{dinu2024symbolicai} positions LLMs as components that function as semantic parsers. These parsers transform inputs (e.g., natural language prompts) into structured data representations or perform operations guided by semantic intent---the specific meaning and purpose the user intends to convey or achieve.

Type theory \citep{andrews2013introduction, church1940formulation, pierce2002types, coquand1986analysis}, particularly through the lens of the Curry-Howard correspondence \citep{wadler2015propositions, dinu2024parameter}, establishes an isomorphism where types in a formal system are analogous to propositions in logic, and well-typed programs (or terms inhabiting those types) correspond to constructive proofs of those propositions \citep{girard1989proofs}. A "well-typed" program or data structure, in this context, is one that conforms to the structural and constraint rules defined by its type. Consequently, defining contracts over such well-typed data structures provides a rigorous theoretical basis for specifying and verifying semantic requirements; a model output satisfying a contract can be seen as a constructive proof of the specification embodied by that contract.

Contracts specify pre-conditions ($P$) that must hold before a component's execution and post-conditions ($Q$) that are guaranteed upon successful, type-conformant completion. The satisfaction of these contractual obligations by LLMs, which are inherently probabilistic, is therefore also probabilistic. The layer incorporates automated, model-driven remediation attempts to guide the component toward outputs that comply with the contract. This methodology extends classical Hoare logic assertions of the form $\{P\}C\{Q\}$ \citep{hoare1969axiomatic, leung2010program}---which specify that if a pre-condition $P$ holds before the execution of a computational unit $C$, then a post-condition $Q$ will hold upon its termination---by associating the fulfillment of such a triple with a quantifiable success probability, $P_{succ}$.

In the context of this work, an agent $\mathcal{A}$ is a tuple of the form
$$
\mathcal{A} = \langle\, \mathcal{M}, \Pi, \Theta, \mathcal{T}, \mathcal{C}\ \rangle
$$
where $\mathcal{M}$ is a set of generative models (e.g., LLMs) the agent controls, $\Pi$ is the set of instructions the agent must execute, $\Theta$ is the set of hyperparameters governing the agent's behavior (e.g., temperature, iterations, etc.), $\mathcal{T}$ is the set of of types the agent can handle, and $\mathcal{C}$ is the set of contracts the agent must satisfy. The agent's behavior is defined by its ability to generate well-typed outputs according to $\mathcal{T}$, given inputs that are well-typed according to $\mathcal{T}$ and instructions in $\Pi$, while respecting hyperparameters $\Theta$ and satisfying contracts $\mathcal{C}$.

Inspired by observational equivalence \citep{morris1968lambda,hennessy1985algebraic,plotkin1981structural,da1992observational}, agents satisfying the same probabilistic contracts are \emph{functionally equivalent} with respect to those contracts; they differ only in (i) $P_{succ}$, (ii) operational costs, and (iii) \emph{potential}---i.e., the capacity of the agent to satisfy an ever more complex set of conditions, enabling principled comparison and selection.

In practice, guardrails and schema validators \citep{nvidia2023nemo, pydantic2025}, function-calling and constitutional-style self-remediation \citep{lin2024hammer, chen2024iteralign, bai2022constitutional,madaan2023self}, and ReAct/RAG correction loops \citep{yao2023react, asai2024self, shinn2023reflexion, yao2023tree, besta2024graph} can all be expressed as pre/post predicates with bounded remediation under the contract layer. Our contribution is a unified, declarative DbC formalism and execution flow that subsumes these patterns while remaining implementation-agnostic.

\section{Related Work}
\textbf{Program Correctness and DbC:} The axiomatic approach \citep{floyd1993assigning, hoare1969axiomatic} and DbC \citep{meyer1992applying} provide foundations for reasoning about software behavior through explicit interface specifications \citep{leung2010program}. While runtime monitoring imposes overhead, soft contract verification \citep{nguyen2014soft} enables static proofs via symbolic execution, and \citet{hanus2020combining} combines static/dynamic checking for compile-time verification.

\textbf{Type Theory:} From Church's simple types \citep{church1940formulation} to Intuitionistic Type Theory \citep{coquand1986analysis, pierce2002types}, type systems ensure well-structured data for contracts. The Curry-Howard correspondence \citep{wadler2015propositions, andrews2013introduction} links types to logical propositions, making type conformance a prerequisite for semantic validation.

\textbf{Probabilistic Program Logics:} Probabilistic Hoare Logics \citep{kozen1979semantics, den2002verifying} establish properties with probabilities, extended by loops \citep{sun2024relative} and union bounds \citep{barthe2016program}. eRHL \citep{avanzini2025quantitative} provides quantitative reasoning for probabilistic programs with completeness results.

\textbf{LLM Reliability:} LLMs produce inconsistent outputs \citep{yan2024large} despite code generation \citep{chen2021evaluating} and semantic parsing capabilities \citep{dinu2024symbolicai, schneider2024evaluating}. PEIRCE \citep{quan2025peirce} unifies material/formal inference through neuro-symbolic conjecture-criticism, while contracts address DL API reliability \citep{ahmed2023design}. Echoing \citet{marcus2001algebraic}, reliability may hinge on architectures that explicitly manipulate symbols, providing the deterministic, compositional substrate today’s LLMs lack.

\section{Implementation}

The contract layer is built entirely on the SymbolicAI framework \citep{dinu2024symbolicai} , extending its neurosymbolic capabilities with DbC-inspired validation mechanisms\footnote{%
  The contract source code is available at this GitHub \href{https://github.com/ExtensityAI/symbolicai/blob/main/symai/strategy.py\#L372}{[link]}; the documentation is available at this GitBook \href{https://extensityai.gitbook.io/symbolicai/features/contracts}{[link]}. For challenging practical use cases involving contracts, see this codebase \href{https://github.com/ExtensityAI/ontology-hydra/tree/main}{[link]}, which includes various types of contracts ranging from simple to complex (e.g., repairing a broken ontology).%
}. At its core, the implementation leverages user-defined data models (Pydantic-based \citep{pydantic2025} \texttt{LLMDataModel} subclasses) that serve as the concrete instantiation of the type system $\mathcal{T}$. These models define not only structural requirements but also rich semantic descriptions through field annotations, guiding both type validation and LLM generation.

\begin{figure}[ht]
    \centering
    \includegraphics[width=0.8\columnwidth]{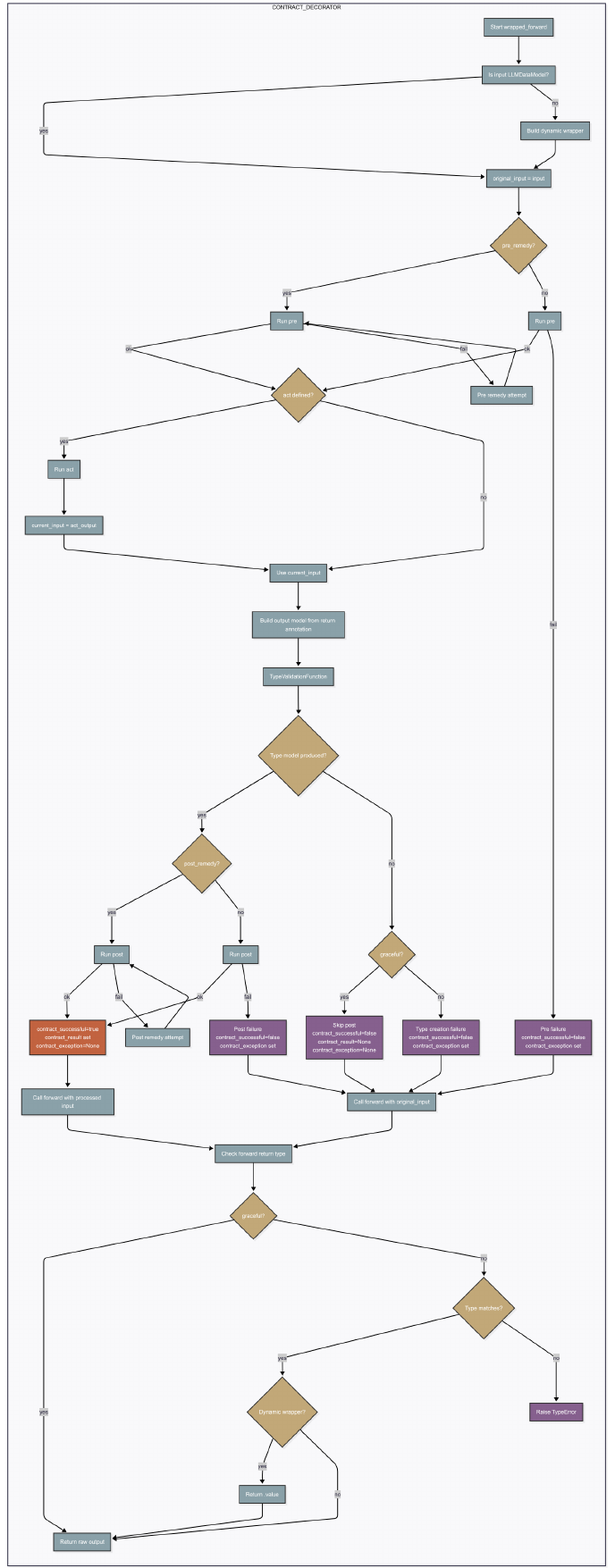}
    \caption{Contract execution flow.}
    \label{fig:flow}
\end{figure}

When an agent $\mathcal{A}$ processes a request, the contract decorator wraps the underlying \texttt{Expression} class's \texttt{forward} method, establishing a validation pipeline. Pre-conditions are implemented as methods that either return \texttt{True} or raise descriptive exceptions. These exceptions serve a dual purpose: they signal contract violations and provide semantic guidance to the LLM during the remediation phase. Similarly, post-conditions validate the generated output's semantic properties beyond mere type conformance.

The contract execution flow proceeds through distinct phases: (i) initial type validation of inputs against $\mathcal{T}$, (ii) pre-condition checking with optional remediation, (iii) an optional \texttt{act} method for intermediate transformations, (iv) LLM-driven output generation guided by the class's prompt property and type specifications, and (v) post-condition validation with remediation attempts. Type validation enforces structural well-formedness, while pre/post enforce semantic predicates on well-typed objects. When remediation is enabled for either pre- or post-conditions, both input and output objects are "fixed"—guaranteed to be type-valid with all field conditions satisfied. This fixing process uses the LLM to populate or correct fields according to the type schema and semantic instructions, ensuring that each field contains valid data (e.g., an extracted email or default value) that passes all specified constraints.

Remediation is achieved through a component that iteratively refines outputs by incorporating validation error messages into corrective prompts. The accumulation of error history across retry attempts enables the LLM to learn from previous failures, avoiding cyclic errors. Each validation function is parameterized by retry configurations in $\Theta$, controlling the maximum attempts (to preclude deep loops), delays, and backoff strategies.

A critical aspect of the implementation is the fallback mechanism, which embodies DbC's invariant principle. The original \texttt{forward} method is always executed in a \texttt{finally} block, regardless of contract validation outcomes. If contract validation fails, the \texttt{forward} method can enable graceful falling, where contract or type mismatches do not raise—the raw \texttt{forward} output is returned as‑is, or can return a safe, type-compliant default. This design guarantees that contract failures never prevent system operation, merely degrading from verified to best-effort behavior.

Lastly, the success of contract satisfaction for each invocation is a Bernoulli random variable. We compute the empirical success probability over N independent runs as
$P_{\text{succ}}=\tfrac{1}{N}\sum_{t=1}^{N}\mathbf{1}[\text{all contract predicates pass}]$. To simplify evaluation and reporting, we approximate the overall success probability by multiplying per-family success probabilities. For example, if a post-condition involves six individual checks (predicates), we can group them into families—such as class existence, index validity, and cluster reduction—and treat these families as independent, even though dependencies may exist. Combined with performance metrics that track validation overhead and remediation costs, this approximation enables runtime comparison of functionally equivalent agents that satisfy the same contracts $\mathcal{C}$.

\section{Limitations and Future Work}
While our contract layer provides a robust framework for ensuring type conformance and semantic validation, several limitations merit discussion, alongside promising avenues for future investigation.

\

\textbf{Model Constraints.} Semantic validation remains bounded by LLM capabilities and stochasticity. While hyperparameter control enables deterministic outputs, low-temperature settings may prune valid solution paths. Recent work on grammar-constrained generation offers promising mitigation strategies \citep{willard2023efficient, geng2023grammar, park2025flexibleefficientgrammarconstraineddecoding}. Libraries like Lark \citep{lark2024} and Parsimonious \citep{parsimonious2022} provide practical foundations for encoding deterministic constraints within our semantic validation framework.

\textbf{Design Trade-offs.} Proactive contract design demands upfront investment. Poorly specified contracts risk over-constraining agents, creating brittle systems that fail to generalize. Conversely, overly permissive contracts provide weak guarantees. The challenge lies in systematically designing constraints that guide generation toward intended outcomes without stifling valid solution paths.

\textbf{Formal Verification.} Current contracts lack formal guarantees about type system correctness. Future work will pursue Lean4 \citep{moura2021lean} formalization of the entire pipeline, providing machine-checked proofs of type safety and contract satisfaction properties. This formalization will establish rigorous foundations for trustworthy agent design, enabling verification of contract consistency, type safety preservation across agent compositions, and probabilistic bounds on contract satisfaction under various operational conditions. The Lean4 effort is prospective; the runtime layer does not depend on it.

\section{Conclusion}

We have presented a DbC inspired layer for trustworthy agent design that bridges the gap between LLM capabilities and formal verification requirements. By extending classical DbC principles to the probabilistic domain of generative models, our approach provides verifiable guarantees through type-theoretic contracts while maintaining the flexibility inherent in LLM-based systems.

\bibliography{main}

\end{document}